\documentclass[runningheads]{llncs}
\usepackage{graphicx}
\usepackage{multirow}
\usepackage{tikz}
\usepackage{subfig}
\usepackage{comment}
\usepackage{amsmath,amssymb}
\usepackage{color}
\usepackage{graphicx}
\usepackage{booktabs}
\usepackage[accsupp]{axessibility}

\begin{document}
\pagestyle{headings}
\mainmatter
\def\ECCVSubNumber{3475}  % Insert your submission number here

\title{Homogeneous Multi-modal Feature Fusion and Interaction for 3D Object Detection} % Replace with your title

% INITIAL SUBMISSION 
\begin{comment}
\titlerunning{ECCV-22 submission ID \ECCVSubNumber} 
\authorrunning{ECCV-22 submission ID \ECCVSubNumber} 
\author{Anonymous ECCV submission}
\institute{Paper ID \ECCVSubNumber}
\end{comment}
%******************

\titlerunning{HMFI}
\author{Xin Li\inst{1} \and
Botian Shi\inst{2} \and
Yuenan Hou\inst{2}\and
Xingjiao Wu\inst{1,3}\and
Tianlong Ma\inst{1,3}\and
\\
Yikang Li\inst{2}$^{\dag}$ \and
Liang He\inst{1}$^{\dag}$ 
}
\authorrunning{X. Li et al.}
\institute{$^1$East China Normal University \quad $^2$Shanghai AI Lab \quad $^3$Fudan University \\
{\tt\small \email{\{sankin0528, wuxingjiao2885\}@gmail.com} \\ 
\tt\small\email{\{shibotian, houyuenan, liyikang\}@pjlab.org.cn}
\tt\small\email{\{tlma, lhe\}@cs.ecnu.edu.cn}}
\\
$^{\dag}$Corresponding author}

\maketitle
\def\algorithmname{HMFI}
\begin{abstract}
Multi-modal 3D object detection has been an active research topic in autonomous driving. Nevertheless, it is non-trivial to explore the cross-modal feature fusion between sparse 3D points and dense 2D pixels. 
Recent approaches either fuse the image features with the point cloud features that are projected onto the 2D image plane or combine the sparse point cloud with dense image pixels. These fusion approaches often suffer from severe information loss, thus causing sub-optimal performance. 
To address these problems, we construct the homogeneous structure between the point cloud and images to avoid projective information loss by transforming the camera features into the LiDAR 3D space. In this paper, we propose a homogeneous multi-modal feature fusion and interaction method (HMFI) for 3D object detection. 
Specifically, we first design an image voxel lifter module (IVLM) to lift 2D image features into the 3D space and generate homogeneous image voxel features. Then, we fuse the voxelized point cloud features with the image features from different regions by introducing the self-attention based query fusion mechanism (QFM). Next, we propose a voxel feature interaction module (VFIM) to enforce the consistency of semantic information from identical objects in the homogeneous point cloud and image voxel representations, which can provide object-level alignment guidance for cross-modal feature fusion and strengthen the discriminative ability in complex backgrounds.
We conduct extensive experiments on the KITTI and Waymo Open Dataset, and the proposed HMFI achieves better performance compared with the state-of-the-art multi-modal methods. Particularly, for the 3D detection of cyclist on the KITTI benchmark, HMFI surpasses all the published algorithms by a large margin.
\end{abstract}

\section{Introduction}
\begin{figure}[t]
\centering
\includegraphics[width=1\textwidth]{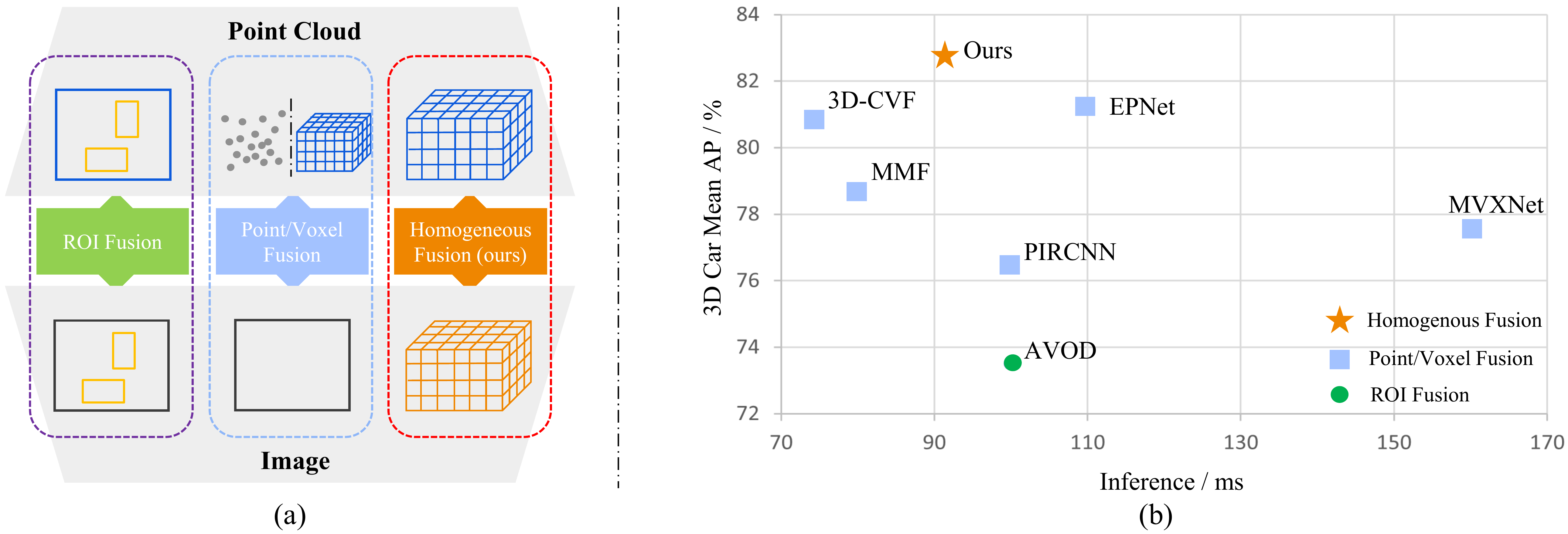}
  \caption{(a) Schematic comparison between different feature-level fusion based methods. (b) Quantitative comparison with competitive multi-modal feature-level fusion methods. Our method achieves good performance-efficiency trade-off for the car category (Mean AP of all difficulty levels) on the KITTI~\cite{kitti} benchmark.}
\label{fig:first}
\end{figure}

3D object detection is an important task that aims to precisely localize and classify each object in the 3D space, thus allowing vehicles to perceive and understand their surrounding environment comprehensively. So far, various LiDAR-based and image-based 3D detection approaches ~\cite{pointnet,pointnet++,caddn,liu2020smoke,pointrcnn,pointpillars,parta2,pvrcnn,liga,voxelrcnn,pyramidrcnn} have been proposed.

LiDAR-based methods can achieve superior performance over image-based approaches as point cloud contains precise spatial information. However, LiDAR points are usually sparse and do not have enough color and texture information. As to image-based approaches, they perform better in capturing semantic information while suffering from the lack of depth signal. Therefore, multi-modal 3D object detection is a promising direction that can fully utilize the complementary information of images and point cloud.

Recent multi-modal approaches can be generally categorized into two types: decision-level fusion and feature-level fusion. Decision-level fusion methods ensemble the detected objects in respective modalities and their performance is bounded by each stage~\cite{clocs}. 
Feature level fusion is more prevalent as they fuse the rich informative features of two modalities. Three representative feature-level fusion methods are depicted in Fig.~\ref{fig:first}(a). The first one is fusing multi-modal features at the regions of interest (RoI). However, these methods have severe spatial information loss when projecting 3D points onto the bird’s eye view (BEV) or front view (FV) in 2D plane, while 3D information plays a key role in accurate 3D object localization. Another line of work conducts fusion on the point/voxel-level~\cite{mvx,pointpainting,pircnn,mmf,confuse,pointaugmenting,epnet,3dcvf}, which can achieve complementary fusion at a much finer granularity and involve the combination of low-level multi-modal features at 3D points or 2D pixels. However, they can only approximately establish a relatively coarse correspondence between the point/voxel features and image features. Moreover, these two schemes of feature fusion usually suffer from severe information loss due to the mismatched projection between 2D dense image pixels and 3D sparse LiDAR points.

To address the aforementioned problems, we propose a homogeneous fusion scheme that lifts image features from 2D plane to 3D dense voxel structure. In our homogeneous fusion scheme, we propose the Homogeneous Multi-modal Feature Fusion and Interaction method (HMFI), which exploits the complementary information in multi-modal features and alleviates severe information loss caused by the dimensional reduction mapping. Furthermore, we build the cross-modal feature interaction between the point cloud features and image features at object-level based on the homogeneous 3D structure to strengthen the model's ability to fuse image semantic information with the point cloud.

Specifically, we design an image voxel lifter module (IVLM) to lift the 2D image features to the 3D space first and construct a homogeneous voxel structure of 2D images for multi-modal feature fusion, which is guided by the point cloud as depth hint. It will not cause information loss for fusing these two multi-modal data. We also notice that the homogeneous voxel structure of cross-modal data can help in feature fusion and interaction. Thus, we introduce the query fusion mechanism (QFM) that introduces a self-attention based operation that can adaptively combine point cloud and image features. Each point cloud voxel will query 
all image voxels to achieve homogeneous feature fusion and combine with the original point cloud voxel features to form the joint camera-LiDAR features. QFM enables each point cloud voxel to perceive image features in the common 3D space adaptively and fuse these two homogeneous representations effectively.

Besides, we explore building a feature interaction between the homogeneous point cloud and image voxel features instead of refining in regions of interest (RoI) based pooling which is applied to fuse low-level LiDAR and camera features with the joint camera-LiDAR features. We consider that, although point cloud and image representations are in different modalities, the object-level semantic properties should be similar in the homogeneous structure. Therefore, to strengthen the abstract representation of point cloud and images in a shared 3D space and exploit the similarity of identical objects' properties in two modalities, we propose a voxel feature interaction module (VFIM) at the object-level to improve the consistency of point cloud and image homogeneous representations in the 3D RoI. To be specific, we use the voxel RoI pooling~\cite{voxelrcnn} to extract features in these two homogeneous features according to the predicted proposals and produce the paired RoI feature set. Then we adopt the cosine similarity loss~\cite{simsiam} between each pair of RoI features and enforce the consistency of object-level properties in point cloud and images. In VFIM, building the feature interaction in these homogeneous paired RoI features improves the object-level semantic consistency between two homogeneous representations and enhances the model’s ability to achieve cross-modal feature fusion. Extensive experiments conducted on KITTI and Waymo Open Dataset demonstrate that the proposed method can achieve better performance compared to the state-of-the-art multi-modal methods. Our \textbf{contributions} are summarized as below:

\begin{enumerate}
    \item 
    We propose an image voxel lifter module (IVLM) to lift 2D image features into the 3D space and construct two homogeneous features for multi-modal fusion, which retains original information of image and point cloud.

    \item We introduce the query fusion mechanism (QFM) to fuse two homogeneous representations of the point cloud voxel features and image voxel features effectively, which enables the fused voxels to perceive objects in a unified 3D space for each frame adaptively.

    \item We propose a voxel feature interaction module (VFIM) to improve the consistency of identical objects' semantic information in the homogeneous point cloud and image voxel features which can guide the cross-modal feature fusion and greatly improve the detection performance.

    \item Extensive experiments demonstrate the effectiveness of the proposed HMFI and achieve competitive performance on KITTI and Waymo Open Dataset. Notably, on the KITTI benchmark, HMFI surpasses all the published competitive methods by a large margin on detecting cyclist.
\end{enumerate}

\section{Related Works}
\subsection{LiDAR-based 3D Object Detection}
\textbf{Point-based methods:} These methods~\cite{pointnet,pointnet++,pointgnn,pointrcnn} take the raw point cloud as input and employ stacked MLP layers to extract point features. PointRCNN~\cite{pointrcnn} uses the PointNets~\cite{pointnet,pointnet++} as point cloud encoder, then generates proposals based on the extracted semantic and geometric features, and refines these coarse proposals via 3D ROI pooling operation. Point-GNN ~\cite{pointgnn} designs a graph neural network to detect 3D objects and encodes the point clouds in a fixed radius near the neighbors' graph. Since the point clouds are unordered and large in number, point-based methods typically suffer from high computational costs.

\noindent\textbf{Voxel-based methods:} These voxel-based approaches~\cite{second,voxelnet,voxelrcnn,parta2,lidarrcnn,pvrcnn,votr} tend to convert the point cloud into voxels
and utilize voxel encoding layers to extract voxel features. SECOND~\cite{second} proposes
a novel sparse convolution layer to replace the original computation-intensive 3D convolution. PointPillars~\cite{pointpillars} converts the point cloud to a pseudo-image and applies 2D CNN to produce the final detection results. Some other works ~\cite{voxelrcnn,pvrcnn,lidarrcnn,parta2,pyramidrcnn} follow~\cite{second} to utilize the 3D sparse convolutional operations to encode the voxel features and obtain
more accurate detection results in the coarse-to-refine two-stage manner.
The more recent CT3D~\cite{ct3d} designs a channel-wise transformer architecture to constitute 3D object detection framework with minimal hand-crafted design.

\subsection{Image-based 3D Object Detection}
Many researchers are also very concerned about how to use camera images to perform 3D detection ~\cite{liu2020smoke,pseudo++,lu2021geometry,liga,caddn}. Specifically, CaDDN~\cite{caddn} designs a Frustum Feature Network to project image information into 3D space. We directly introduce depth bins through point cloud projection and use a non-parametrical module to lift image features into 3D space. LIGA-Stereo~\cite{liga} utilizes the LiDAR-based model to guide the training of stereo-based 3D detection model and achieves the state-of-the-art stereo detection performance. Although cameras are the most common sensors and inexpensive, the performance of image-based methods is still inferior to the LiDAR-based approaches due to the lack of accurate depth information.

\subsection{Multi-modal 3D Object Detection}
Multi-modal 3D object detection has received more and more attention~\cite{survey2021} as it can utilize the complementary information of each single modality to the maximum extent. There are two levels of fusion: decision-level
fusion~\cite{avod,mv3d,qi2018frustum,frustumconv,clocs}and feature-level fusion~\cite{pointfusion,3dcvf,pointpainting,pircnn,epnet,mvx,mmf,confuse}. The former fusion methods~\cite{clocs} ensemble the detection results of each modality directly. Their performance is limited by each stage. 
% For example, CLOCs~\cite{clocs} encodes the detection results from images and point cloud separately, and then follows a coarse-to-refine stage to refine these results. 

As for feature-level fusion methods which fuse multi-modal data in a much finer granularity, AVOD~\cite{avod} utilizes point clouds BEV as well as images features and feeds the features into region proposal network (RPN) for improving detection performance. F-ConvNet~\cite{frustumconv} follows ~\cite{qi2018frustum} to utilize frustum point clouds and front view images for 3D object detection. PointFusion~\cite{pointfusion} and PointPainting~\cite{pointpainting}
enhance raw point cloud with the corresponding class prediction scores through a well pre-trained image semantic segmentation network~\cite{maskrcnn}. 
EPNet~\cite{epnet} projects the point cloud into image plane to retrieve semantic information at multi-level resolutions in a point-wise manner.
MVXNet~\cite{mvx} utilizes pre-trained 2D detectors~\cite{fasterrcnn} to produce semantic image features to strengthen the voxel feature representations in the early stage. These methods only exploit part of the rich
information contained in an image and suffer from severe information loss~\cite{survey}. 3D-CVF~\cite{3dcvf} lifts image features to the dense 3D voxel space but fuses the multi-modal feature in BEV via a cross-view spatial feature fusion strategy and it causes feature overlap in 3D space when constructing image voxel features.

Although many multi-modal networks have been proposed, they do not easily outperform state-of-the-art LiDAR-only based detectors. These fusion methods establish a coarse relationship between the point cloud features and semantic image features. Besides, they suffer from severe information loss by perspective projection. Moreover, existing fusion methods do not exploit the similarity of object-level semantic information in the cross-modal fusion. Our approach is designed to overcome these challenges and achieve better 3D detection performance.
\begin{figure}[t]
\centering
\includegraphics[width=1\textwidth]{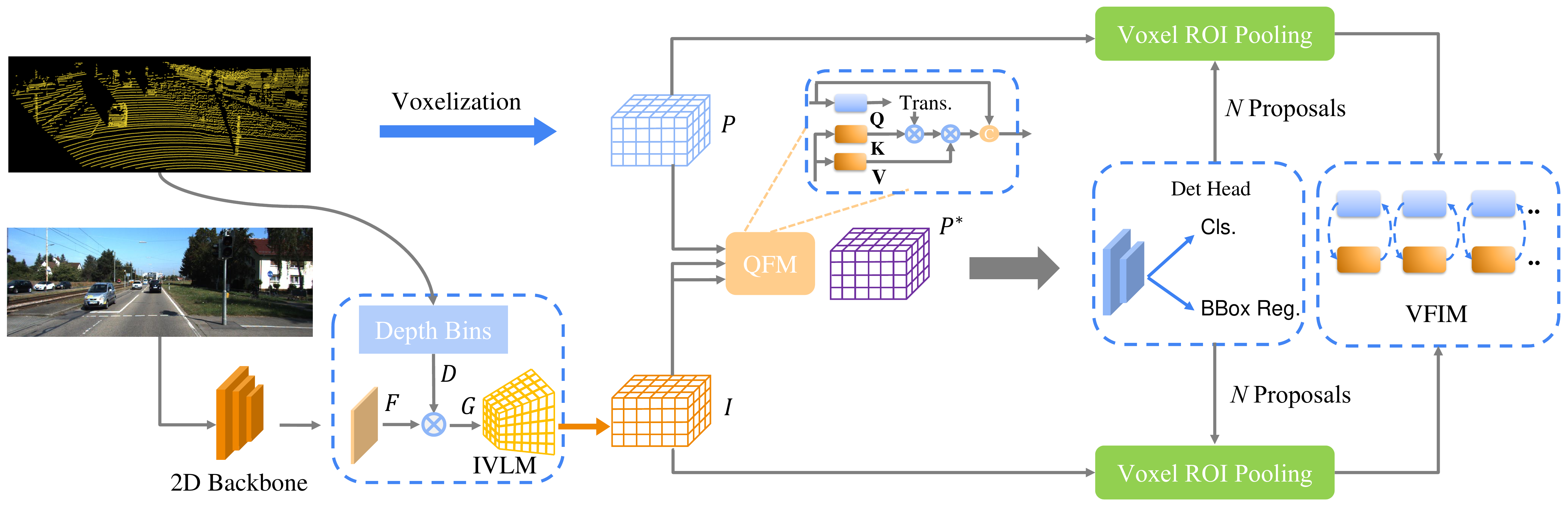}
  \caption{The architecture of HMFI. Each image is processed by a 2D backbone network and fed into an image voxel lifter module (IVLM) to produce a homogeneous structure based on the depth bins transformed by the point cloud. Then, the processed homogeneous image and point cloud features are fused by the query fusion mechanism (QFM). Next, a voxel-based object detector is employed on fused features to produce 3D detection results. Finally, the voxel feature interaction module (VFIM) conducts feature interaction at object-level based on the detection results to improve semantic consistency in these two homogeneous cross-modal features.}
\label{fig:architecture}
\end{figure}

\subsection{Methodology}
\subsection{Framework Overview}
The overall architecture of the proposed homogeneous multi-modal fusion and interaction (HMFI) method is illustrated in Fig. \ref{fig:architecture}. We first leverage a point encoding network to extract the features of the point cloud and then pool them to obtain the voxel features $P\in\mathbb{R}^{X_P\times Y_P\times Z_P\times C_F}$~\cite{voxelnet} where the $C_F$ is the number of channels of the voxel feature and the ($X_P, Y_P, Z_P$) is the grid size. The image $\tilde I\in \mathbb R ^{{W_{\tilde I}} \times {H_{\tilde I}} \times 3}$ is fed into a ResNet-50~\cite{resnet50} backbone to extract image features $F\in \mathbb R ^{{W_F} \times {H_F} \times C_F}$, where $W_{\tilde I}$ and $H_{\tilde I}$ are the width and height of the image and $W_F$, $H_F$ and $C_F$ are the width, height and number of channels of the image features. 

In order to fuse point cloud features and image features in 3D space, we propose an image voxel lifter module (IVLM) to project the image feature $F$ into 3D homogeneous image voxel space as the $I\in\mathbb{R}^{X_I\times Y_I\times Z_I\times C_F}$. Then we use the query fusion mechanism (QFM) to fuse the homogeneous point voxel $P$ and image voxel $I$ to generate the fused representation $P^*\in\mathbb{R}^{X_P\times Y_P\times Z_P\times C_F}$. Afterward, we use the detection module to generate the classification and 3D bounding box of each object based on $P^*$. Meanwhile, a voxel feature interaction module (VFIM) is proposed to conduct the feature interaction at object-level based on the detection results to improve semantic consistency in these two homogeneous cross-modal features. We introduce the details in the following sections.

\subsection{Image Voxel Lifter Module}
\begin{figure}[t]
\centering
\includegraphics[width=1\textwidth]{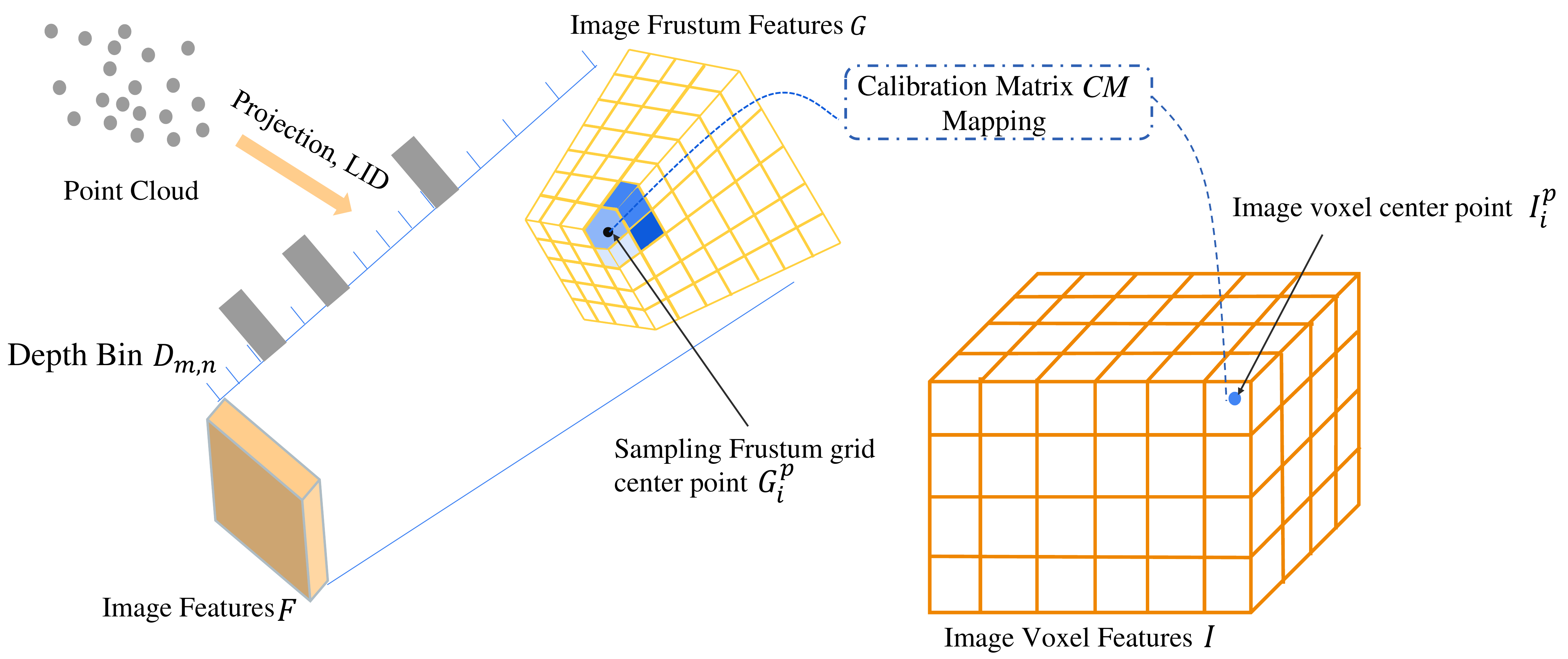}
  \caption{The image voxel lifter module. Each feature pixel $F_{m, n}$ along the ray is determined by the discrete depth bins $D$ to generate frustum features $G_{m, n}$. Then sampling grid center points in image voxel are projected into the frustum grid based on the calibration matrix $CM$. The neighboring sampled voxel grids (shown as blue in the image frustum features $G$) are combined using trilinear interpolation and assigned to the corresponding voxel in $I$.}
\label{fig:ivlm}
\end{figure}

To encode perceptual depth information in the image effectively and construct a homogeneous structure for multi-modal feature fusion and interaction, we propose the image voxel lifter module (IVLM) to lift 2D image features into 3D space by associating image features and discretized depth maps. The procedure is shown in Fig.~\ref{fig:ivlm}.

To construct an image feature voxel, we follow~\cite{lifteccv2020,caddn} and convert the image plane features into frustum features $G$ which can
encode depth information in image features.  Thus, we scatter the vector $F_{m, n}\in \mathbb{R}^{C_F}$ of each pixel $(m, n)$ in the image feature map $F$ into the 3D space determined by the depth bin $D_{m, n}$ along the ray of image frustum perspective projection.
 
The depth bins $D$ are produced by discretizing the depth map with a linear-increasing depth discretization (LID) method \cite{center3d,caddn}. The $D\in\mathbb{R}^{W_F\times H_F\times R}$ consists of $W_F\times H_F$ one-hot discretized depth bins in $\mathbb{R}^R$. In order to associate image features with discretized depth information, we utilize the outer product to process the image features $F$ and depth bins $D$ to generate a frustum feature $G\in\mathbb{R}^{W_F\times H_F\times R\times C_F}$. Each $G_{m,n}\in\mathbb{R}^{R\times C_F}$ on pixel $(m,n)$ can be calculated by:
\begin{equation}
G_{m,n} = F_{m,n}\otimes D_{m,n}
\label{eq:generateG}
\end{equation}
where $\otimes$ represents the outer product, $(m, n)$ is the index of the each feature pixel.

Next, we transform the features from the frustum space $G\in\mathbb{R}^{W_F\times H_F\times R\times C_F}$ into the 3D space $I\in\mathbb{R}^{X_I\times Y_I\times Z_I\times C_F}$ by the trilinear interpolation. 

Specifically, to acquire the $i$-th image voxel feature $I_i\in\mathbb{R}^{C_F}$, we sample the corresponding centroid in image frustum features $G$ by a transformation based on calibration matrix $CM$ as $G_i^p=CM\cdot I_i^p$, where the $G_i^p, I_i^p\in\mathbb{R}^3$ indicates the 3D position of the $i$-th grid in $G$, $I$. After that, we conduct the trilinear interpolation around the neighborhood of $G_i^p$ to form the $I_i^p$. Finally, the image voxel features $I$ is constructed by this process on each spatial index $i$.

\subsection{Query Fusion Mechanism}
To exploit the complementary information from point cloud and images, we introduce the query fusion mechanism (QFM) that enables each point cloud voxel feature to perceive the whole image and selectively combines image voxel features. Instead of simply fusing the cross-modal voxel pairs, we consider that the LiDAR voxel can perceive the whole image voxel feature. In order to aggregate these complementary information of two modalities effectively, we propose to use a self-attention~\cite{transformer} module which regards each voxel feature vector of image and point cloud as a homogeneous token. 

To be more specific, we use the point cloud voxel features $F_{P}$ as the queries, the image voxel features $F_{I}$ as the keys as well as values to conduct the fusion and form the fused voxel features $P^*$. The construction of $F_{P}$ and $F_{I}$ is described as follows.

Considering that most of LiDAR voxels are empty, we produce $F_{P}\in\mathbb{R}^{M\times C_F}$ by selecting all $M$ non-empty voxels within the homogeneous point cloud voxel features $P$. However, the image voxel features $I$ is much denser than point cloud voxels. In order to reduce the computational cost, we adopt the 3D max-pooling on $I$ with a scale factor $\lambda$ to obtain the most informative features $I^*\in\mathbb{R}^{\frac{X_I}{\lambda}\times\frac{Y_I}{\lambda}\times\frac{Z_I}{\lambda}\times C_F}$. Then, we flatten $I^*$ along the first three dimensions to make $F_{I}\in\mathbb{R}^{L\times C_F}$ where $L=\frac{X_I}{\lambda}*\frac{Y_I}{\lambda}*\frac{Z_I}{\lambda}$.

After constructing the point voxel $F_{P}$ and the image voxel $F_{I}$, we utilize a multi-head self-attention\cite{transformer} layer as the query fusion mechanism (QFM). We adopt three learnable linear transformation for each head $i$ on the query $F_{P}$, key $F_{I}$ and value $F_{I}$, denoted as $Q_i\in\mathbb{R}^{M\times d_k}$, $K_i\in\mathbb{R}^{L\times d_k}$ and $V_i\in\mathbb{R}^{L\times d_v}$ respectively:
\begin{align}
Q_i=F_{P}\cdot W_i^Q,~~~~
K_i=F_{I}\cdot W_i^K,~~~~
V_i=F_{I}\cdot W_i^V
\end{align}
where $W_i^Q\in\mathbb{R}^{C_F\times d_k}$, $W_i^K\in\mathbb{R}^{C_F\times d_k}$ and $W_i^V\in\mathbb{R}^{C_F\times d_v}$.

Then we perform the multi-head self-attention with $r$ heads:
\begin{align}
\begin{array}{c}
A_M = \text{Concat}(\text{head}_1, \text{head}_2, \cdots, \text{head}_r)W^O \\
\text{head}_i=\text{softmax}\left(\frac{Q_i K_i^T}{\sqrt{d_k}}\right)V_i
\end{array}
\end{align}
where $A_M\in\mathbb{R}^{M\times C_F}$ is the output of multi-head attention module, $W_O\in\mathbb{R}^{r*d_v\times C_F}$ is a linear transformation matrix to project the concatenation of $r$ attention heads into the homogeneous point voxel space. Then we concatenate the $A_M$ and the non-empty point voxel features $F_{P}$ to acquire the fused voxel features $F_{P}^*\in\mathbb{R}^{M\times (2*C_F)}$. Finally, we restore $F_{P}^*$ into the homogeneous voxel space as $P^*\in\mathbb{R}^{X_P\times Y_P\times Z_P\times (2*C_F)}$ as the input of the downstream 3D object detection module.

\begin{figure}[t]
\centering
\includegraphics[width=1\textwidth]{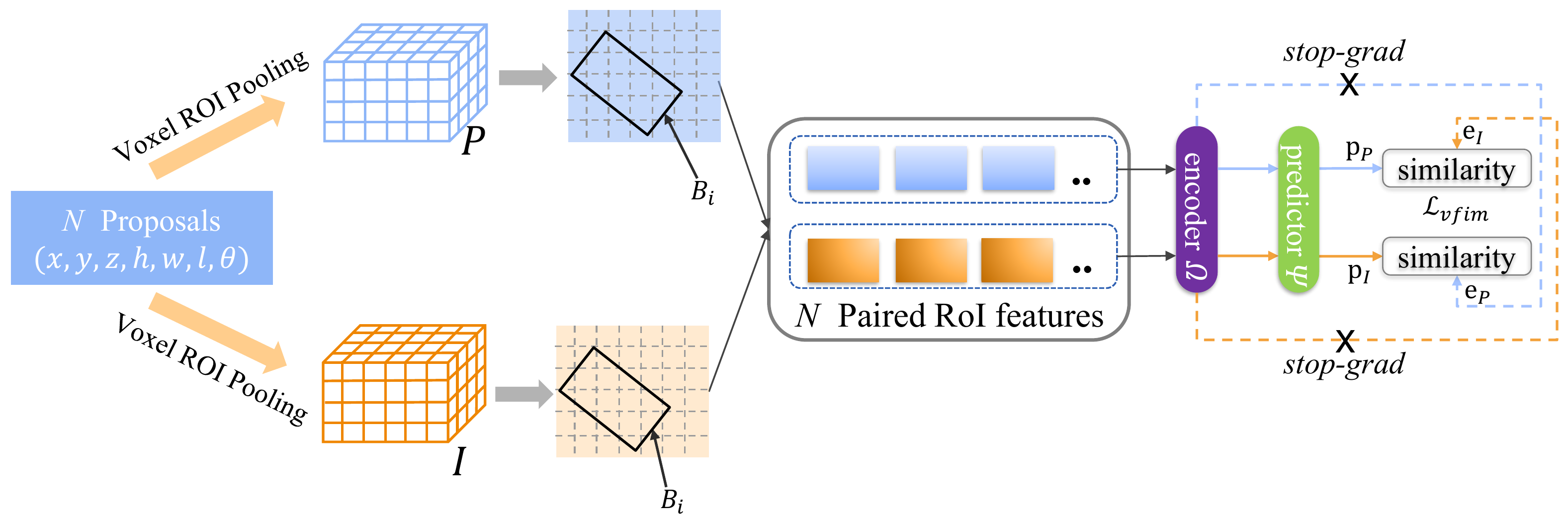}
  \caption{The voxel feature interaction module. We use the voxel RoI pooling to extract features in these two homogeneous features according to the predicted proposals and form the paired RoI feature set. Then we adopt feature interaction between each pair of RoI features based on the symmetry similarity constraint loss to improve the object-level semantic consistency between two homogeneous representations}
\label{fig:vfim}
\end{figure}
\subsection{Voxel Feature Interaction Module}
LiDARs and cameras have different representations of identical objects in the scene. Though the modalities are different from each other, the object-level representations should be similar. 
Motivated by this observation, we design a voxel feature interaction module (VFIM) to build the feature interaction in these two cross-modal features based on the consistency of object-level properties in point cloud and images. And we can fully utilize the similarity constraint between homogeneous features $P$ and $I$ with the object-level guidance to achieve better cross-modal feature fusion. 

As shown in Fig.~\ref{fig:vfim}, we sample the $N$ 3D detection proposals from 3D detection head as $B=\{B_1, B_2, ,..., B_N\}$. Then, we introduce the voxel RoI pooling~\cite{voxelrcnn} on homogeneous point voxel features $P$ and image voxel features $I$ to obtain the respective RoI features including $P_B=\{P_{B_1},P_{B_2},\dots,P_{B_N}\}$ and $I_B=\{I_{B_1},I_{B_2},\dots,I_{B_N}\}$. 

Finally, inspired by~\cite{simsiam}, to improve the similarity between the output vectors from the paired RoI features $P_{B_i}$ and $I_{B_i}$, we feed both of them into an encoder $\Omega$ and use a MLP-based predictor $\Psi$ to transform the output of these encoded RoI features into the metric space: $p_P=\Psi(\Omega(P_{B_i}))$, $e_I=\Omega(I_{B_i})$, $p_I=\Psi(\Omega(I_{B_i}))$, $e_P=\Omega(P_{B_i})$.

We minimize the paired feature distance by using cosine similarity:
% and match them to each other. 
% Denoting the two output vectors as $p_1 = mp(enc(P_{B_i}))$ and $e_1 = enc(IV_{B_i})$, we minimize their feature distance by cosine similarity:
\begin{equation}
\begin{array}{c}
CosSim(p,e) =  - \frac{{{p}}}{{{{\left\| {{p}} \right\|}_2}}}*\frac{{{e}}}{{{{\left\| {{e}} \right\|}_2}}}
\end{array}
\label{eq:losscossim}
\end{equation}
where ${\left\| \cdot \right\|}_2$ means $l_2$ normalization.

Meanwhile, the stop-gradient operation is also adopted for better modeling the similarity constraint, then we utilize the symmetry similarity constraint loss $\mathcal{L}_{vfim}$  as:
\begin{equation}
\begin{array}{c}
\mathcal{L}_{vfim} = \frac{1}{2}CosSim(p_P,stop\_grad(e_I)) + \frac{1}{2}CosSim(p_I,stop\_grad(e_P))
\end{array}
\label{eq:lossvfim}
\end{equation}

\subsection{Loss Function}
In previous methods, the image backbone is directly initialized with the fixed pre-trained weights from other external datasets such as ImageNet. On the contrary, our HMFI is trained via two-stage training process in an end-to-end manner. We utilize a multi-task loss function for jointly optimizing the whole network. The total loss $\mathcal{L}_{total}$ can be formulated as:
\begin{equation}
\begin{array}{c}
{\mathcal{L}_{{\rm{total}}}}{\rm{ = }}{\mathcal{L}_{{\rm{rpn}}}} + {\mathcal{L}_{rcnn}}+\gamma \mathcal{L}_{vfim}
\end{array}
\label{eq:losstotal}
\end{equation}
where $\gamma$ is set to 0.1. $\mathcal{L}_{rpn}$ and $\mathcal{L}_{rcnn}$ denote the training objectives for the region proposal network (RPN) and the refinement network,  
We follow ~\cite{voxelrcnn,second} to devise the loss of the RPN as:
\begin{equation}
% \begin{array}{c}
{\mathcal{L}_{{\rm{rpn}}}}{\rm{ = }}{\omega _1}{\mathcal{L}_{{\rm{cls}}}} + {\omega _2}{\mathcal{L}_{reg}}
% \end{array}
\label{eq:lossrpn}
\end{equation}
where $\omega_1$ and $\omega_2$ are set to 1 and 2, 
respectively.  We adopt the focal loss~\cite{focalloss} to balance the positive and negative
samples in classification loss with default hyperparameters and the $smooth_{L1}$ loss is utilized for the box regression.
The proposal refinement loss $\mathcal{L}_{rcnn}$ includes the IoU-guided confidence prediction loss $\mathcal{L}_{iou}$ and the box refinement loss $\mathcal{L}_{refine}$ as
\begin{equation}
{\mathcal{L}_{{\rm{rcnn}}}}{\rm{ = }}{\mathcal{L}_{{\rm{iou}}}} + {\mathcal{L}_{refine}}
\label{eq:lossrcnn}
\end{equation}

\section{Experiments}
In this section, we evaluate the performance of the proposed HMFI on the KITTI~\cite{kitti} and Waymo Open Dataset~\cite{waymo}.

\subsection{Datasets}
\noindent\textbf{KITTI} is a widely used dataset. It consists of 7,481 training frames and 7,518 testing frames, with 2D and 3D annotations of cars, pedestrians and cyclists on the streets. Objects are divided into three difficulty levels: easy, moderate and hard, according to their size, occlusion level and truncation level. For validation, training samples are commonly divided into a train set with 3,712 samples and a val set with 3,769 samples.

\noindent\textbf{Waymo Open Dataset} (WOD) is a large-scale dataset for autonomous driving.
There are totally 798 scenes for training and 202 scenes for validation. Each scene is a sequential segment that has around 20 seconds of sensor data. Note that cameras in WOD only cover around $250^\circ$ field of view (FOV), which is different from LiDAR points and 3D labels in full $360^\circ$. To follow the same setting of KITTI, we only select LiDAR points and ground-truth in the FOV of front camera for training and evaluating. We sample every 5$^{th}$ frames from all the training samples to form the new training set ($\sim$32k frames) due to the large dataset size and high frame rate.

\subsection{Implementation Details}
\noindent\textbf{Experimental Settings.} On the KITTI benchmark, we set the range of point cloud to [0, 70.4], [-40, 40], [-3, 1]m in the (x, y, z) axis.  The LiDAR voxel structure is divided by a voxel size (0.05, 0.05, 0.1)m, while each image voxel size is set to (0.2, 0.2, 0.4)m to fit with the feature size of the point cloud branch. As for Waymo, we use [0, 75.2], [-75.2, 75.2], [-2, 4]m for the point cloud range, (0.1, 0.1, 0.15)m for the voxel size. And each image voxel size is set to (0.4, 0.4, 0.6)m to fit the point cloud feature size. In the QFM, the scale factor $\lambda$ is set as 4, the count $r$ and hidden units of attention heads are set to 4 and 64, respectively. In the VFIM, the settings of the voxel RoI pooling operation are the same as Voxel-RCNN~\cite{voxelrcnn} and we sample $N =$ 128 proposals, half of them are positive samples that have $IoU > 0.55$ with the corresponding ground truth boxes. The number of hidden units of the encoder $\Omega$ and predictor $\Psi$ are both set to 256. 

\begin{table}[htbp]
\caption{Quantitative comparison with the state-of-the-art 3D object detection methods on KITTI \textit{test} set.}
\centering
\scalebox{0.78}{
\begin{tabular}{c|c|ccc|ccc|ccc}
\hline\hline                                &                                     & \multicolumn{3}{c|}{Car 3D AP}  & \multicolumn{3}{c|}{Pedestrian 3D AP}& \multicolumn{3}{c}{Cyclist 3D AP}        \\
\multirow{-2}{*}{{ Method}} & \multirow{-2}{*}{Modality}   & Easy        & Mod.       & Hard      & Easy         & Mod.        & Hard & Easy         & Mod.        & Hard   \\ \hline
              PointPillars~\cite{pointpillars}                                   &    LiDAR                       &  82.58& 74.31 &68.99 &51.45 &41.92&38.89&77.10 &   58.65 &   51.92                        \\
              SECOND~\cite{second}                                  &       LiDAR                      &       87.44 &79.46 &73.97                  &   -          &     -       &         -    &   -          &     -       &         -   \\
             
              PointRCNN~\cite{pointrcnn}                         &  LiDAR                          &   86.96 &75.64& 70.70 &47.98 &39.37&36.01&          74.96& 58.82& 52.53                           \\ 
              PointGNN~\cite{pointgnn}                                  &  LiDAR                           & 88.33& 79.47& 72.29 &51.92&\textbf{43.77}&40.14                      &         78.60& 63.48& 57.08       \\ 
              Part A$^2$~\cite{parta2}                          &  LiDAR                           &        87.81 &78.49& 73.51
              &\textbf{53.10}&43.35&40.06&79.17& 63.52& 56.93       \\ 
              PV-RCNN~\cite{pvrcnn}                                 &   LiDAR                          &  90.25 & 81.43 &76.82 &52.17&43.29&\textbf{40.29}                   &78.60& 63.71& 57.65       \\
              Voxel-RCNN~\cite{voxelrcnn}                                  &  LiDAR                          &\textbf{90.90}& 81.62 &77.06                       &   -          &      -      &        -  &   -          &      -      &        - \\

              M3DETR~\cite{m3detr}                                 &   LiDAR                    &90.28& 81.73& 76.96 &   -          &      -      &        -& \textbf{83.83}& \textbf{66.74} & \textbf{59.03}     \\  
              CT3D~\cite{ct3d}                                 &   LiDAR                          &   87.83 &81.77 & 77.16                     &      -       &      -      &      - &      -       &      -      &      -     \\ 
              Pyramid RCNN-V~\cite{pyramidrcnn}                                 &   LiDAR                          &    87.06 & 81.28& 76.85                &      -       &      -      &      -   &      -       &      -      &      - \\ 
              Pyramid RCNN-PV~\cite{pyramidrcnn}                                 &   LiDAR                          &     88.39 & \textbf{82.08} & \textbf{77.49}	                    &      -       &      -      &      -   &      -       &      -      &      - \\ \hline
              MV3D~\cite{mv3d}                                  &      LiDAR+RGB                      &  74.97 &63.63 &54.00                       &  -           &       -     &        -      &      -       &      -      &      - \\
              Confuse~\cite{voxelrcnn}                                 &        LiDAR+RGB                    &                        83.68& 68.78 &61.67 &      -   &           -   &     -    &      -       &      -      &      -      \\
              F-PointNet~\cite{qi2018frustum}                                 &        LiDAR+RGB                    &                        82.19& 69.79&60.59 &      50.53          &   42.15  &    38.08    &  72.27      &     56.12&49.01    \\
              MVXNet~\cite{mvx}                                  &     LiDAR+RGB                       &   83.20  & 72.70  & 65.20  &- &    -      &     - &      -       &      -      &      -  \\
              PointPainting~\cite{pointpainting}                                  & LiDAR+RGB                           &     82.11 &71.70 &67.08   &   50.32      &      40.97     &     37.77     &     77.63  & 63.78 & 55.89             \\
              AVOD-FPN~\cite{avod}                                  &      LiDAR+RGB                      & 83.07 &71.76 & 65.73&      50.46    &      42.27    &     39.04   & 63.76 & 50.55& 44.93         \\ 
              MAFF~\cite{maff} &   LiDAR+RGB           &85.52   &75.04 &  67.60&-&-&-&      -       &      -      &      - \\
              PI-RCNN~\cite{pircnn}                                 &        LiDAR+RGB                    &    84.37 &74.82 &70.03                      &       -      &      -      &        -  &      -       &      -      &      -           \\
              F-Convnet~\cite{frustumconv}                                &   LiDAR+RGB                         &  87.36 & 76.39 &66.69         &     \textbf{52.16}     &  \textbf{43.38}      &    38.80      &     81.98 &65.07 &56.54               \\ 
              MMF~\cite{mmf}                                 &        LiDAR+RGB                    &     88.40 & 77.43 & 70.22                    &    -         &       -     &        -      &      -       &      -      &      -       \\
              CLOCs\_PVCas ~\cite{clocs}                                 &   LiDAR+RGB                         &     88.94 & 80.67 &  77.15                    &   47.30          &     39.42   &   36.97    &     77.33      &      62.02    &    55.52         \\
              EPNet~\cite{epnet}                                   &        LiDAR+RGB                    &     \textbf{89.81} & 79.28 & 74.59                    &      -       &     -     &        -  &      -       &      -      &      -          \\
              3D-CVF~\cite{3dcvf}                                 &       LiDAR+RGB                     &     89.20  & 80.05 & 73.11                    &        -     &     -       &       -    &      -       &      -      &      -             \\ \hline
              HMFI (ours)                                &          LiDAR+RGB                  &     88.90                    &  \textbf{81.93}&   \textbf{77.30} &      50.88    &    42.65   &     \textbf{39.78}     &   \textbf{84.02}         &  \textbf{70.37}            & \textbf{62.57}            \\ 
\hline\hline
\end{tabular}
\label{tab:kittitest}
}
\end{table}

\noindent\textbf{Training.} To validate the effectiveness of our HMFI, we select the Voxel-RCNN~\cite{voxelrcnn} as the baseline. Our HMFI is trained via the two-stage training process. We adopt OpenPCDet~\cite{openpcdet} as our codebase, and a pre-trained ResNet50\cite{resnet50} is adopted as the 2D backbone to produce image features $F$ for the image voxel lifter module. We train the model with the Adam~\cite{adam} optimizer, which uses the one-cycle policy ~\cite{learnrate} with the initial learning rate being 0.0005. Batch size is set as 2. The total number of training epochs is set as 80 for KITTI~\cite{kitti} and 30 epochs for WOD~\cite{waymo}.

\subsection{Results on KITTI dataset}
\textbf{KITTI Test set.} Experiments on the KITTI test split~\cite{kitti} are evaluated using average precision (AP) via 40 recall positions. We compare our HMFI with other state-of-the-art approaches by submitting the detection results to the KITTI server for evaluation. Table~\ref{tab:kittitest} presents the quantitative comparison with state-of-the-art 3D object detection methods on the KITTI test set. It is apparent that the HMFI achieves better or comparable performance over the state-of-the-art methods on car and cyclist for all difficulty levels, respectively. The HMFI achieves up to 1.88\% gains (for moderate difficulty) over 3D-CVF~\cite{3dcvf} which is the best feature-level fusion based method. The HMFI outperforms most of the LiDAR-based 3D object detectors except for the Pyramid RCNN-PV~\cite{pyramidrcnn} which introduces the raw point features to achieve a better result but with a worse efficiency. By contrast, our method outperforms the Pyramid RCNN-V~\cite{pyramidrcnn} in the same settings. Especially, our HMFI surpasses all the published algorithms by a large margin for the 3D detection of cyclist. Note that none of the models in Table~\ref{tab:kittitest} can achieve superior performance to our model on car and cyclist simultaneously.

\noindent\textbf{KITTI Val set.} In addition, we also report
the performance on the KITTI val set with AP calculated by
11 recall positions. As shown in Table~\ref{table:val}, our HMFI achieves the state-of-the-art performance on moderate level on the val set, even better than the LiDAR-based method~\cite{pyramidrcnn}. 

To sum up, the results on both val set and test set consistently demonstrate that our proposed HMFI achieves superior 3D detection performance. Specifically, we achieve satisfactory performance on pedestrian and cyclist which usually have very few points in LiDAR measurements. As shown in Fig.~\ref{fig:first} (b), we also report the inference time per frame of some feature-level fusion methods, and our HMFI achieves the best balance between the accuracy and efficiency among all methods.
\begin{table}[htbp]
\centering
\caption{Performance comparison on the moderate level of KITTI val split with AP calculated by 11 recall positions, $\dag$ means the our re-implementation results. Car$_{Mod.}$, Pedestrian$_{Mod.}$ and Cyclist$_{Mod.}$ donate the performance of Car, Pedestrian and Cyclist on moderate level respectively.  }

\begin{tabular}{c|c|c|c|c}
\hline
Method&Modality & Car$_{Mod.}$ &Pedestrian$_{Mod.}$  & Cyclist$_{Mod.}$                                 \\\hline
SECOND~\cite{second} &LiDAR &    76.48   &    59.84  & 64.89                \\
PointRCNN~\cite{pointrcnn}&LiDAR & 76.05 &51.59&	66.67\\
PV-RCNN~\cite{pvrcnn}  &LiDAR & 83.69 & 57.90& 70.47\\
Voxel-RCNN~\cite{voxelrcnn}  &LiDAR  &84.52 &- &-        \\
Voxel-RCNN$^\dag$~\cite{voxelrcnn}  &LiDAR  &84.27 &60.11 &72.07        \\
Pyramid RCNN-PV~\cite{pyramidrcnn}  &LiDAR & 84.38& -  &    -               \\\hline
PointPainting~\cite{pointrcnn} &LiDAR+RGB& 77.74 &61.67 &	71.62\\
CLOCs\_SecCas~\cite{clocs}&LiDAR+RGB&79.31&56.20&67.92\\
3D-CVF~\cite{3dcvf} &LiDAR+RGB& 79.88 &-	&-\\\hline
 HMFI (ours)  &LiDAR+RGB & \textbf{85.14}   & \textbf{62.41} & \textbf{74.11} \\ \hline
\end{tabular}
\label{table:val}
\end{table}

\subsection{Results on Waymo Open Dataset}
To further validate the effectiveness of the proposed HMFI, we also conduct experiments on the large-scale Waymo Open Dataset. 
Two difficulty levels are also introduced, where the LEVEL\_1 mAP is calculated on objects that have more than 5 points and the LEVEL\_2 mAP is measured on objects that have 1$\sim$5 points. 
Table~\ref{tab:waymo} summarizes the performance of our method and baselines. It is obvious that our HMFI performs superbly over all the object classes and two difficulty levels. In particular, we achieve remarkable gains on pedestrian and cyclist with +2.17\% and +1.86\% mAP on LEVEL 2, which demonstrates the outstanding performance of our method on detecting objects with fewer than 5 LiDAR points. The results on the Waymo Open Dataset further validate both the effectiveness and generalization of the HMFI.
\begin{table}[htbp]
\caption{Performance comparison on the Waymo Open Dataset with 202 validation sequences ($\sim$40k samples)}
\centering
\scalebox{0.88}{\begin{tabular}{c|cc|cc|cc}
\hline
\multirow{2}{*}{Method}& \multicolumn{2}{c|}{Vehicle L1/L2} & \multicolumn{2}{c|}{Pedestrian L1/L2} & \multicolumn{2}{c}{Cyclist L1/L2} \\ 
                       &AP              & APH             & AP               & APH               & AP              & APH            \\ \hline
    Baseline       &         66.46/64.21        &     64.81/62.40         &               64.35/62.74        &        58.11/55.43           &    62.34/59.85      &           59.64/57.67                \\
    Ours        &      68.34/65.66        &     66.84/64.57           &       66.62/64.91           &     59.76/57.24              &       64.25/61.71         &     61.23/59.21           \\ \hline
  Improvements   &   +1.88/+1.45 & +2.03/+2.17             &    +2.27/+2.17 &  +1.65/+1.81         &   +1.91/+1.86              &  +1.59/+1.54                              \\ \hline

\end{tabular}}
\label{tab:waymo}
\end{table}

\subsection{Ablation Study}
In this section, we present an ablation study for validating the effect of each component in the HMFI method. The ablation study is conducted on the KITTI validation set. We adopt the mean average precision (mAP) on easy, moderate and hard difficulty levels via 11 recall positions for evaluation. As shown in Table~\ref{table:ablationstudy}, our HMFI can bring over 1.8\% AP performance gain on all difficulty levels of three objects.

\noindent\textbf{Effect of Query Fusion Mechanism.} The query fusion mechanism  (QFM) combines the image and point cloud features selectively depending on their relevance according to the attention map between the image features and point cloud features. In Table~\ref{table:ablationstudy}, we observe that QFM can generate the enhanced joint camera-LiDAR features and lead to 0.83\%, 0.58\%, and 0.62\% performance gains in AP$_{Easy}$, AP$_{Mod.}$, AP$_{Hard}$, respectively.
\begin{table}[ht]
\caption{Effect of each component of our HMFI on KITTI val set.
$AP_{Easy}$, $AP_{Mod.}$, and $AP_{Hard} $ are the mAP performance of easy, moderate, and hard levels respectively.}
\centering
\begin{tabular}{cccc|c|c|c}
\hline
Method                & QFM & IVLM & VFIM & \begin{tabular}[c]{@{}c@{}} AP$_{Easy}$\end{tabular} & \begin{tabular}[c]{@{}c@{}}AP$_{Mod.}$\end{tabular} & \begin{tabular}[c]{@{}c@{}}AP$_{Hard}$\end{tabular} \\ \hline
Baseline~\cite{voxelrcnn}             &   -   &   -  &  -    &       81.34 &	71.76 &	67.09  \\ \hline
\multirow{3}{*}{Ours} &  \checkmark    &  -   &   -   &    82.17 &	72.34 & 67.73\\
                      &   \checkmark    &  \checkmark    &  -    &   82.52 &	72.94 &	68.45 \\
                      &   \checkmark   & \checkmark    &   \checkmark   &83.36&	73.89 & 68.98  \\ \hline
 \multicolumn{4}{c|}{Improvements} & +2.02 & +2.13 & +1.89  \\
                     \hline
\end{tabular}
\label{table:ablationstudy}
\end{table}

\noindent\textbf{Effect of Multi-modal Feature Structure.} In Table~\ref{table:ablationstudy}, We observe that the IVLM can bring 0.35\%, 0.60\%, and 0.72\% performance gains in AP$_{Easy}$, AP$_{Mod.}$, AP$_{Hard}$. IVLM lifts image features to the homogeneous space with point cloud voxel features, which not only facilitates feature fusion, but also enables object-level semantic consistency modeling between two homogeneous features.

\noindent\textbf{Effect of Voxel Feature Interaction.}
We observe that the voxel feature interaction module (VFIM) improves the baseline by 0.84\%, 0.95\%, and 0.53\% in AP$_{Easy}$, AP$_{Mod.}$, AP$_{Hard}$, respectively. It indicates that our VFIM plays a pivotal role in our multi-modal detection framework. It can improve object-level semantic consistency between two homogeneous features and enables the detector to aggregate paired features across homogenous representations based on object-level semantic similarity.

\section{Conclusions}
In this paper, we propose the homogeneous multi-modal feature fusion and interaction (HMFI) method for 3D detection which fuses image and point cloud features in a homogeneous structure and enforces the consistency of object-level semantic information between two homogeneous features. 
We propose an image voxel lifter module (IVLM) to lift 2D image features to the 3D space and generate homogeneous image voxel features with point cloud voxel features. Then, image and point cloud features are selectively combined by the query fusion mechanism (QFM). Besides, we build the feature interaction in the homogeneous image and point cloud voxel features based on the similarity of object-level semantic information. Extensive experiments conducted on KITTI and Waymo Open Dataset show that significant performance gains can be obtained by our proposed HMFI. Particularly, for the detection of cyclist on the KITTI benchmark, HMFI surpasses all published algorithms by a large margin.
\\
\\
\noindent{ \bf Acknowledgments.} This research is funded by the Science and Technology Commission of Shanghai Municipality (19511120200), The computation is performed in ECNU Multifunctional Platform for Innovation (001).

\bibliographystyle{splncs04}
% \bibliography{egbib}

\begin{thebibliography}{10}
\providecommand{\url}[1]{\texttt{#1}}
\providecommand{\urlprefix}{URL }
\providecommand{\doi}[1]{https://doi.org/#1}

\bibitem{fog}
Bijelic, M., Gruber, T., Mannan, F., Kraus, F., Ritter, W., Dietmayer, K.,
  Heide, F.: Seeing through fog without seeing fog: Deep multimodal sensor
  fusion in unseen adverse weather. In: CVPR. pp. 11682--11692 (2020)

\bibitem{chen20153d}
Chen, X., Kundu, K., Zhu, Y., Berneshawi, A.G., Ma, H., Fidler, S., Urtasun,
  R.: 3d object proposals for accurate object class detection. In: NeurIPS. pp.
  424--432. Citeseer (2015)

\bibitem{chen20173d}
Chen, X., Kundu, K., Zhu, Y., Ma, H., Fidler, S., Urtasun, R.: 3d object
  proposals using stereo imagery for accurate object class detection. IEEE
  TPAMI  \textbf{40}(5),  1259--1272 (2017)

\bibitem{mv3d}
Chen, X., Ma, H., Wan, J., Li, B., Xia, T.: Multi-view 3d object detection
  network for autonomous driving. In: CVPR. pp. 1907--1915 (2017)

\bibitem{simsiam}
Chen, X., He, K.: Exploring simple siamese representation learning. In:
  Proceedings of the IEEE/CVF Conference on Computer Vision and Pattern
  Recognition. pp. 15750--15758 (2021)

\bibitem{voxelrcnn}
Deng, J., Shi, S., Li, P., Zhou, W., Zhang, Y., Li, H.: Voxel r-cnn: Towards
  high performance voxel-based 3d object detection. In: AAAI. pp. 1201--1209
  (2021)

\bibitem{kitti}
Geiger, A., Lenz, P., Urtasun, R.: Are we ready for autonomous driving? the
  kitti vision benchmark suite. In: CVPR. pp. 3354--3361 (2012)

\bibitem{m3detr}
Guan, T., Wang, J., Lan, S., Chandra, R., Wu, Z., Davis, L., Manocha, D.:
  M3detr: Multi-representation, multi-scale, mutual-relation 3d object
  detection with transformers. In: WACV. pp. 772--782 (2022)

\bibitem{liga}
Guo, X., Shi, S., Wang, X., Li, H.: Liga-stereo: Learning lidar geometry aware
  representations for stereo-based 3d detector. In: CVPR. pp. 3153--3163 (2021)

\bibitem{maskrcnn}
He, K., Gkioxari, G., Doll{\'a}r, P., Girshick, R.: Mask r-cnn. In: ICCV. pp.
  2961--2969 (2017)

\bibitem{resnet101}
He, K., Zhang, X., Ren, S., Sun, J.: Deep residual learning for image
  recognition. In: CVPR. pp. 770--778 (2016)

\bibitem{resnet50}
He, K., Zhang, X., Ren, S., Sun, J.: Deep residual learning for image
  recognition. In: Proceedings of the IEEE conference on computer vision and
  pattern recognition. pp. 770--778 (2016)

\bibitem{mono3d++}
He, T., Soatto, S.: {Mono3d++: Monocular 3d vehicle detection with two-scale 3d
  hypotheses and task priors}. In: AAAI. vol.~33, pp. 8409--8416 (2019)

\bibitem{epnet}
Huang, T., Liu, Z., Chen, X., Bai, X.: Epnet: Enhancing point features with
  image semantics for 3d object detection. In: ECCV. pp. 35--52. Springer
  (2020)

\bibitem{adam}
Kingma, D.P., Ba, J.: Adam: A method for stochastic optimization. ICLR  (2015)

\bibitem{fuseseg}
Krispel, G., Opitz, M., Waltner, G., Possegger, H., Bischof, H.: Fuseseg: Lidar
  point cloud segmentation fusing multi-modal data. In: WACV. pp. 1874--1883
  (2020)

\bibitem{avod}
Ku, J., Mozifian, M., Lee, J., Harakeh, A., Waslander, S.L.: Joint 3d proposal
  generation and object detection from view aggregation. In: IROS. pp.~1--8.
  IEEE (2018)

\bibitem{pointpillars}
Lang, A.H., Vora, S., Caesar, H., Zhou, L., Yang, J., Beijbom, O.:
  Pointpillars: Fast encoders for object detection from point clouds. In: CVPR.
  pp. 12697--12705 (2019)

\bibitem{li2019stereo}
Li, P., Chen, X., Shen, S.: Stereo r-cnn based 3d object detection for
  autonomous driving. In: CVPR. pp. 7644--7652 (2019)

\bibitem{lidarrcnn}
Li, Z., Wang, F., Wang, N.: Lidar r-cnn: An efficient and universal 3d object
  detector. In: CVPR. pp. 7546--7555 (2021)

\bibitem{mmf}
Liang, M., Yang, B., Chen, Y., Hu, R., Urtasun, R.: Multi-task multi-sensor
  fusion for 3d object detection. In: CVPR. pp. 7345--7353 (2019)

\bibitem{confuse}
Liang, M., Yang, B., Wang, S., Urtasun, R.: Deep continuous fusion for
  multi-sensor 3d object detection. In: ECCV. pp. 641--656 (2018)

\bibitem{focalloss}
Lin, T.Y., Goyal, P., Girshick, R., He, K., Doll{\'a}r, P.: Focal loss for
  dense object detection. In: ICCV. pp. 2980--2988 (2017)

\bibitem{liu2020smoke}
Liu, Z., Wu, Z., T{\'o}th, R.: Smoke: Single-stage monocular 3d object
  detection via keypoint estimation. In: CVPRW. pp. 996--997 (2020)

\bibitem{lu2021geometry}
Lu, Y., Ma, X., Yang, L., Zhang, T., Liu, Y., Chu, Q., Yan, J., Ouyang, W.:
  Geometry uncertainty projection network for monocular 3d object detection.
  In: ICCV. pp. 3111--3121 (2021)

\bibitem{pyramidrcnn}
Mao, J., Niu, M., Bai, H., Liang, X., Xu, H., Xu, C.: Pyramid r-cnn: Towards
  better performance and adaptability for 3d object detection. In: Proceedings
  of the IEEE/CVF International Conference on Computer Vision. pp. 2723--2732
  (2021)

\bibitem{votr}
Mao, J., Xue, Y., Niu, M., Bai, H., Feng, J., Liang, X., Xu, H., Xu, C.: Voxel
  transformer for 3d object detection. In: ICCV. pp. 3164--3173 (2021)

\bibitem{sensorfusion}
Meyer, G.P., Charland, J., Hegde, D., Laddha, A., Vallespi-Gonzalez, C.: Sensor
  fusion for joint 3d object detection and semantic segmentation. In:
  Proceedings of the IEEE/CVF Conference on Computer Vision and Pattern
  Recognition Workshops. pp.~0--0 (2019)

\bibitem{clocs}
Pang, S., Morris, D., Radha, H.: Clocs: Camera-lidar object candidates fusion
  for 3d object detection. In: IROS. pp. 10386--10393. IEEE (2020)

\bibitem{lifteccv2020}
Philion, J., Fidler, S.: Lift, splat, shoot: Encoding images from arbitrary
  camera rigs by implicitly unprojecting to 3d. In: European Conference on
  Computer Vision. pp. 194--210. Springer (2020)

\bibitem{pon2020object}
Pon, A.D., Ku, J., Li, C., Waslander, S.L.: Object-centric stereo matching for
  3d object detection. In: ICRA. pp. 8383--8389 (2020)

\bibitem{qi2018frustum}
Qi, C.R., Liu, W., Wu, C., Su, H., Guibas, L.J.: Frustum pointnets for 3d
  object detection from rgb-d data. In: CVPR. pp. 918--927 (2018)

\bibitem{pointnet}
Qi, C.R., Su, H., Mo, K., Guibas, L.J.: Pointnet: Deep learning on point sets
  for 3d classification and segmentation. In: CVPR. pp. 652--660 (2017)

\bibitem{pointnet++}
Qi, C.R., Yi, L., Su, H., Guibas, L.J.: Pointnet++: Deep hierarchical feature
  learning on point sets in a metric space. NeurIPS  \textbf{30} (2017)

\bibitem{qin2019triangulation}
Qin, Z., Wang, J., Lu, Y.: Triangulation learning network: from monocular to
  stereo 3d object detection. In: CVPR. pp. 7615--7623 (2019)

\bibitem{caddn}
Reading, C., Harakeh, A., Chae, J., Waslander, S.L.: Categorical depth
  distribution network for monocular 3d object detection. In: CVPR. pp.
  8555--8564 (2021)

\bibitem{fasterrcnn}
Ren, S., He, K., Girshick, R., Sun, J.: Faster r-cnn: Towards real-time object
  detection with region proposal networks. NIPS  \textbf{28} (2015)

\bibitem{ct3d}
Sheng, H., Cai, S., Liu, Y., Deng, B., Huang, J., Hua, X.S., Zhao, M.J.:
  Improving 3d object detection with channel-wise transformer. In: ICCV. pp.
  2743--2752 (2021)

\bibitem{pvrcnn}
Shi, S., Guo, C., Jiang, L., Wang, Z., Shi, J., Wang, X., Li, H.: Pv-rcnn:
  Point-voxel feature set abstraction for 3d object detection. In: CVPR. pp.
  10529--10538 (2020)

\bibitem{pointrcnn}
Shi, S., Wang, X., Li, H.: Pointrcnn: 3d object proposal generation and
  detection from point cloud. In: CVPR. pp. 770--779 (2019)

\bibitem{parta2}
Shi, S., Wang, Z., Shi, J., Wang, X., Li, H.: From points to parts: 3d object
  detection from point cloud with part-aware and part-aggregation network. PAMI
   \textbf{43}(8),  2647--2664 (2020)

\bibitem{pointgnn}
Shi, W., Rajkumar, R.: Point-gnn: Graph neural network for 3d object detection
  in a point cloud. In: CVPR. pp. 1711--1719 (2020)

\bibitem{mvx}
Sindagi, V.A., Zhou, Y., Tuzel, O.: Mvx-net: Multimodal voxelnet for 3d object
  detection. In: 2019 International Conference on Robotics and Automation
  (ICRA). pp. 7276--7282. IEEE (2019)

\bibitem{learnrate}
Smith, L.N.: A disciplined approach to neural network hyper-parameters: Part
  1--learning rate, batch size, momentum, and weight decay. arXiv preprint
  arXiv:1803.09820  (2018)

\bibitem{waymo}
Sun, P., Kretzschmar, H., Dotiwalla, X., Chouard, A., Patnaik, V., Tsui, P.,
  Guo, J., Zhou, Y., Chai, Y., Caine, B., et~al.: Scalability in perception for
  autonomous driving: Waymo open dataset. In: Proceedings of the IEEE/CVF
  conference on computer vision and pattern recognition. pp. 2446--2454 (2020)

\bibitem{center3d}
Tang, Y., Dorn, S., Savani, C.: Center3d: Center-based monocular 3d object
  detection with joint depth understanding. In: DAGM German Conference on
  Pattern Recognition. pp. 289--302. Springer (2020)

\bibitem{openpcdet}
Team, O.D.: Openpcdet: An open-source toolbox for 3d object detection from
  point clouds. \url{https://github.com/open-mmlab/OpenPCDet} (2020)

\bibitem{transformer}
Vaswani, A., Shazeer, N., Parmar, N., Uszkoreit, J., Jones, L., Gomez, A.N.,
  Kaiser, {\L}., Polosukhin, I.: Attention is all you need. NIPS  \textbf{30}
  (2017)

\bibitem{pointpainting}
Vora, S., Lang, A.H., Helou, B., Beijbom, O.: Pointpainting: Sequential fusion
  for 3d object detection. In: CVPR. pp. 4604--4612 (2020)

\bibitem{pointaugmenting}
Wang, C., Ma, C., Zhu, M., Yang, X.: Pointaugmenting: Cross-modal augmentation
  for 3d object detection. In: CVPR. pp. 11794--11803 (2021)

\bibitem{survey}
Wang, Y., Mao, Q., Zhu, H., Zhang, Y., Ji, J., Zhang, Y.: Multi-modal 3d object
  detection in autonomous driving: a survey. arXiv preprint arXiv:2106.12735
  (2021)

\bibitem{survey2021}
Wang, Y., Mao, Q., Zhu, H., Zhang, Y., Ji, J., Zhang, Y.: Multi-modal 3d object
  detection in autonomous driving: a survey. CoRR  (2021)

\bibitem{frustumconv}
Wang, Z., Jia, K.: Frustum convnet: Sliding frustums to aggregate local
  point-wise features for amodal 3d object detection. In: IROS. pp. 1742--1749.
  IEEE (2019)

\bibitem{wang2018fusing}
Wang, Z., Zhan, W., Tomizuka, M.: Fusing bird’s eye view lidar point cloud
  and front view camera image for 3d object detection. In: IV. pp.~1--6 (2018)

\bibitem{pircnn}
Xie, L., Xiang, C., Yu, Z., Xu, G., Yang, Z., Cai, D., He, X.: Pi-rcnn: An
  efficient multi-sensor 3d object detector with point-based attentive
  cont-conv fusion module. In: AAAI. pp. 12460--12467 (2020)

\bibitem{pointfusion}
Xu, D., Anguelov, D., Jain, A.: Pointfusion: Deep sensor fusion for 3d bounding
  box estimation. In: CVPR. pp. 244--253 (2018)

\bibitem{second}
Yan, Y., Mao, Y., Li, B.: Second: Sparsely embedded convolutional detection.
  Sensors  \textbf{18}(10), ~3337 (2018)

\bibitem{centerpoint}
Yin, T., Zhou, X., Krahenbuhl, P.: Center-based 3d object detection and
  tracking. In: CVPR. pp. 11784--11793 (2021)

\bibitem{3dcvf}
Yoo, J.H., Kim, Y., Kim, J., Choi, J.W.: 3d-cvf: Generating joint camera and
  lidar features using cross-view spatial feature fusion for 3d object
  detection. In: ECCV. pp. 720--736. Springer (2020)

\bibitem{pseudo++}
You, Y., Wang, Y., Chao, W.L., Garg, D., Pleiss, G., Hariharan, B., Campbell,
  M., Weinberger, K.Q.: Pseudo-lidar++: Accurate depth for 3d object detection
  in autonomous driving. In: ICLR (2020)

\bibitem{moca}
Zhang, W., Wang, Z., Loy, C.C.: Exploring data augmentation for multi-modality
  3d object detection. arXiv preprint arXiv:2012.12741  (2020)

\bibitem{maff}
Zhang, Z., Zhang, M., Liang, Z., Zhao, X., Yang, M., Tan, W., Pu, S.: Maff-net:
  Filter false positive for 3d vehicle detection with multi-modal adaptive
  feature fusion. arXiv preprint arXiv:2009.10945  (2020)

\bibitem{voxelnet}
Zhou, Y., Tuzel, O.: Voxelnet: End-to-end learning for point cloud based 3d
  object detection. In: CVPR. pp. 4490--4499 (2018)

\end{thebibliography}

\end{document}